\documentclass[10pt,twocolumn,letterpaper]{article}

\usepackage{graphicx}
\usepackage{amsmath}
\usepackage{amssymb}
\usepackage{booktabs}
\usepackage{ctable}
\usepackage{multirow}
\usepackage{array}
\usepackage{colortbl}
\usepackage{makecell}
\usepackage{subcaption}
\usepackage{pifont}
\newcommand{\cmark}{\ding{51}}
\newcommand{\xmark}{\ding{55}}

\definecolor{cvprblue}{rgb}{0.21,0.49,0.74}
\usepackage[pagebackref,breaklinks,colorlinks,citecolor=cvprblue]{hyperref}

\graphicspath{{img/}{../img/}{./}{../}}

\begin{document}

\title{Similarity-Guided Curriculum Fine-Tuning of LLMs\\ for Neural Architecture Synthesis}

\author{%
  Anujaya Vijayakumar{\thanks{Corresponding author:
  anujaya.vijayakumar@stud-mail.uni-wuerzburg.de}},
  \hspace{0.2cm} Radu Timofte,
  \hspace{0.2cm} Dmitry Ignatov\\
  \small{Computer Vision Lab, CAIDAS \& IFI,
  University of W\"urzburg, Germany}
}
\date{}
\maketitle

\begin{abstract}

We introduce a MinHash-based similarity scheduling framework that
constructs a progressive curriculum over neural architecture code for
LLM-based neural architecture search (NAS). Using 128-permutation
MinHash signatures over normalised 7-gram source code shingles, we
partition the reference pool into similarity bands and present them in
increasing architectural heterogeneity, with the best LoRA adapter from
each stage merged cumulatively into the backbone. We evaluate the
framework on OlympicCoder-7B within the LEMUR benchmark on CIFAR-10
image classification, generating $N{=}15$ candidate architectures per
epoch across six progressive fine-tuning steps. The curriculum achieves $60\%$ peak success rate at the
high-similarity level without post-processing repair. A $2{\times}2$
ablation at the most diverse level  curriculum versus base model,
with versus without partial interface repair  reveals that without
repair the base model ($47\%$ peak SR) substantially outperforms the
curriculum model ($7\%$ SR), while adding partial repair brings both
to $53\%$ SR. This pattern is consistent with merge-level weight drift
progressively erasing evaluator-interface priors, and suggests that
interface repair and curriculum scheduling target distinct failure modes.
We further report a cross-dataset transfer observation on SVHN, where
direct base-model generation without curriculum warmup yields $27\%$
peak SR at substantially lower accuracy ($60.5\%$) than the CIFAR-10
equivalent, consistent with the increased synthesis difficulty of the
\texttt{unq}-family anchor architecture.

\end{abstract}

\section{Introduction}
\label{sec:intro}

Large language models (LLMs) have emerged as a viable alternative to
classical neural architecture search (NAS) by reframing architecture
design as a program synthesis problem. Instead of searching over
predefined graph templates, an LLM can directly generate executable
Python implementations of complete neural network classes. Frameworks
such as NNGPT~\cite{ABrain.NNGPT} and LEMUR~\cite{ABrain.NN-Dataset,ABrain.LEMUR2}
have demonstrated this feasibility: when conditioned on prior successful
architectures, a code-generating LLM can iteratively propose candidate
models compiled, trained, and scored by an external evaluator.

A question that has received little attention in LLM-based NAS is
\emph{how the reference architectures should be ordered during
fine-tuning}. Existing pipelines treat the reference pool as an
unordered collection. Neural architecture code differs not only in
downstream accuracy but also in module composition, parameter routing,
residual connectivity, and tensor-shape conventions. Exposing all such
structural variations simultaneously may destabilise the internal code
priors the LLM must build to generate evaluator-compatible designs
reliably.

We investigate whether ordering references by structural similarity
during fine-tuning affects generation reliability. We introduce a
MinHash-based similarity scheduling framework that partitions the
reference pool into bands of increasing architectural heterogeneity and
presents them progressively, with sequential LoRA adapter merging to
preserve coding conventions across stages. A $2{\times}2$ ablation
examines how progressive ordering interacts with post-generation
interface repair, revealing that the two components address distinct
failure modes in evaluator-compatible code generation.

This paper makes three contributions:
\begin{enumerate}
    \item A code-level structural similarity pipeline implementing
    MinHash signature computation over normalised 7-gram source
    code shingles, integrated into the LEMUR database, enabling
    sub-second pairwise similarity retrieval across 13,023
    CIFAR-10 neural architecture implementations.
    
    \item A MinHash-based curriculum scheduling framework that
    uses these signatures to partition the reference pool into
    similarity bands of increasing architectural heterogeneity,
    with a systematic viability analysis across six LEMUR datasets.
    
    \item A $2{\times}2$ ablation characterising the interaction
    between interface repair and curriculum warmup in
    evaluator-compatible neural architecture generation.
\end{enumerate}

We note upfront that the single-seed, single-model, single-dataset
experimental scope limits the generality of our conclusions. The
findings reported here are preliminary empirical observations rather
than statistically validated claims, and we treat them as such
throughout.

\section{Related Work}
\label{sec:related}

\subsection{LLM-Based Neural Architecture Search}

Recent work combines LLMs with automated model design in two broad
directions. Optimization by Prompting (OPRO)~\cite{Yang2023OPRO} uses
a frozen LLM as a natural-language optimiser iteratively refining
candidates via scalar feedback; FunSearch~\cite{Romera2023FunSearch}
extends this to program discovery. These approaches use the LLM as a
proposal engine without fine-tuning and do not address reference
scheduling.

EvoPrompting~\cite{Chen2023EvoPrompting} formulates NAS as source-code
generation with the LLM as a mutation operator. ReEvo~\cite{Ye2024ReEvo}
incorporates reflection-based refinement. The NNGPT and LEMUR pipelines
demonstrate that repeated fine-tuning on successful generations gradually
improves architectural quality. LLMatic~\cite{Zheng2023LLMatic} and
GPT-NAS~\cite{Nasir2024GPTNAS} explore structured prompting
strategies. None of these systems explicitly control the structural
difficulty of the supervision signal over time, which is the focus of
this work.

\subsection{Curriculum Learning}

Bengio et al.~\cite{Bengio2009Curriculum} showed that presenting
training samples in meaningful order can improve optimisation and
generalisation. Self-paced learning~\cite{Kumar2010SPL} and
automatic curriculum generation~\cite{Graves2017ACG} followed;
anti-curriculum studies show the benefit of ordering is
task-dependent~\cite{Hacohen2019AntiCurr}.

In code generation, curricula have mostly been applied to
programming-task complexity or staged reasoning supervision. The
present setting differs: difficulty lies in the \emph{architectural
heterogeneity of target code} rather than task complexity, motivating
a similarity-based difficulty measure over source code structure.

\subsection{Code Similarity and Retrieval}

MinHash and Jaccard-based hashing provide efficient large-scale
near-duplicate detection~\cite{Broder1997MinHash}. Alternative
measures include token-level edit distance, AST-based structural
comparison~\cite{Zhang2019ASTDiff}, and neural code embeddings such
as CodeBERT~\cite{Feng2020CodeBERT}. We compare these strategies
in Section~\ref{sec:retrieval} and justify the MinHash choice for the
curriculum context.

\subsection{Parameter-Efficient Continual Tuning}

LoRA-based fine-tuning~\cite{Hu2021LoRA} has become standard for
updating large code models under limited GPU budgets. Sequential
adapter merging, as employed here, is related to model merging
literature  model soups~\cite{Ilharco2023ModelSoups},
TIES-merging~\cite{Yadav2023TIES}, and DARE~\cite{Yu2023DARE}
 but serves curriculum-continuity rather than ensemble combination.

\section{Methodology}

\subsection{Problem Formulation}

Let $\mathcal{D}=\{(c_i,a_i)\}_{i=1}^{|\mathcal{D}|}$ denote the
LEMUR repository, where $c_i$ is the executable Python implementation
of model $i$ and $a_i$ its observed validation accuracy. Given a
pre-trained code LLM $M_0$, we seek an adapted generator $M^\ast$
whose outputs: (1) are accepted by the LEMUR evaluator without
structural failure, and (2) achieve competitive downstream accuracy.

At curriculum stage $t$, a subset $\mathcal{R}_t \subset \mathcal{D}$
of structurally related references is embedded into the prompt context.
The current model $M_t$ generates $N{=}15$ candidates evaluated through
the LEMUR API. Evaluator-valid generations become supervised fine-tuning
examples from which a LoRA adapter $\Delta_t$ is learned. After
convergence, $\Delta_t$ is merged into the backbone to form $M_{t+1}$.

\subsection{MinHash-Based Reference Retrieval}
\label{sec:retrieval}

\paragraph{Retrieval strategy.}
Three reference retrieval approaches are relevant for LLM-based NAS.
\textbf{Accuracy-based retrieval} selects references by performance
rank; the top-$k$ by accuracy may all belong to the same architectural
family, providing no structural diversity signal.
\textbf{Embedding-based retrieval} (e.g., CodeBERT) offers richer
structural discrimination but requires an encoder inference pass per
candidate  computationally prohibitive at the 13,023-architecture
corpus scale across repeated curriculum epochs.
\textbf{MinHash-based retrieval} approximates Jaccard similarity over
normalised 7-gram code token sets using 128-permutation signatures
computed once during corpus preprocessing. 
 
\subsubsection{Signature Computation and Storage}
 
Source code is first tokenised using a regex-based lexer that extracts alphanumeric identifiers and punctuation tokens. From the token sequence, overlapping 7-gram shingles are constructed and encoded as UTF-8 strings. A 128-permutation MinHash signature is computed from these shingles using the \texttt{datasketch} library (\texttt{NNAnalysisSimilarity.py}) and stored in the LEMUR database at model registration time. Each signature is serialised to a compact 512-byte binary BLOB using \texttt{struct.pack(`<128I', *hashvalues)} and written to the \texttt{nn\_minhash} table:
 
\begin{verbatim}
CREATE TABLE nn_minhash (
    nn          TEXT PRIMARY KEY,
    hashvalues  BLOB NOT NULL,   -- 512-byte struct-packed MinHash vector
    num_perm    INTEGER NOT NULL, -- 128
    shingle_n   INTEGER NOT NULL, -- 7
    created_at  TEXT NOT NULL
)
\end{verbatim}

Signatures are batch-upserted at corpus ingestion time via \texttt{upsert\_minhash\_batch()} and individually upserted for each newly evaluated model via \texttt{upsert\_minhash()}. This design means similarity computation incurs zero overhead during fine-tuning epochs, all vectors are pre-computed and stored. The \texttt{nn\_similarity} table defined in the schema is not populated during curriculum runs.So no $O(N^2)$ precomputation of pairwise similarities is performed.

\subsubsection{On-Demand Similarity via SQLite Scalar UDF}
 
A critical architectural decision distinguishes this implementation from the standard \texttt{datasketch} MinHashLSH workflow: \textbf{no Python-side LSH index is constructed at query time}. Instead, Jaccard similarity is computed entirely within the SQLite engine via a registered scalar user-defined function (UDF), \texttt{jaccard\_blobs}, registered on the connection at initialisation.

The UDF takes two stored BLOBs, deserialises each into its 128-permutation hashvalue vector, and returns the MinHash Jaccard estimate as the fraction of matching permutation values:
 
\begin{equation}
\hat{J}(A, B) \;=\; \frac{1}{128} \sum_{i=1}^{128} \mathbf{1}\!\left[h_i(A) = h_i(B)\right]
\label{eq:minhash_udf}
\end{equation}
 
where $h_i(A)$ is the $i$-th permutation hashvalue of architecture $A$, stored as a \texttt{uint32} in the BLOB. The returned value is the MinHash approximation to the true Jaccard coefficient $J(S_A, S_B) = |S_A \cap S_B| / |S_A \cup S_B|$, where $S_A$ and $S_B$ are the 7-gram shingle sets of architectures $A$ and $B$ respectively. \textbf{This is an approximation throughout the pipeline} the 128-permutation estimator has expected variance
\begin{equation}
\mathrm{Var}\!\left[\hat{J}\right] = \frac{J(1 - J)}{128},
\label{eq:minhash_var}
\end{equation}
which is below $0.002$ at all band boundaries used in this work.
 
The Jaccard coefficient over normalised code tokens has a direct structural interpretation: $J = 0.95$ indicates that 95\% of 7-gram token sequences are shared between two implementations, corresponding to near-identical module compositions. This interpretability distinguishes MinHash from learned embedding distances (e.g.\ CodeBERT) for curriculum band construction the band boundaries carry a verifiable structural meaning rather than being defined in an opaque latent space.

\subsubsection{Anchor-Band Retrieval Query}
 
Band-constrained retrieval is implemented in \texttt{\_anchor\_band\_db()} (\texttt{Query.py}) as a single SQL pass. The anchor architecture's BLOB is extracted from a materialised temporary candidate table (\texttt{\_prepare\_anchor\_candidates()}), and the UDF is called per candidate row within a CTE.

The candidate table is a temporary table created per query session, joining \texttt{stat} with \texttt{nn\_minhash} and retaining only the best epoch per unique NN. Band filtering is therefore a SQL predicate rather than a post-processing step, and the SQLite query planner applies it inline during the CTE scan. This architecture achieves sub-second retrieval across the full stored corpus without any Python-side index construction overhead.

\subsection{Dataset viability.}
MinHash curriculum construction requires a structurally dense reference
database with models spanning multiple similarity bands.
Table~\ref{tab:viability} reports this analysis across LEMUR datasets.

\begin{table}[h]
\centering
\fontsize{7.5}{9}\selectfont
\begin{tabular}{l c c c c}
\toprule
\textbf{Dataset} & \textbf{Models} & \textbf{Best acc.} &
\textbf{Max $J$} & \textbf{Viable} \\
\midrule
CIFAR-10      & 13,023 & 0.9717 & 0.98 & \cmark \\
SVHN          &  4,568 & 0.9651 & 0.82$^\ddagger$ & \xmark \\
CelebA-Gender &  3,719 & 0.9847 & 0.99$^\dagger$ & \xmark \\
MNIST         &  3,315 & 0.9976 & 0.23 & \xmark \\
CIFAR-100     &  3,170 & 0.7286 &   & \xmark \\
ImageNette    &  2,804 & 0.9720 & 0.15 & \xmark \\
\bottomrule
\end{tabular}
\caption{Dataset viability for MinHash curriculum construction.
Viable = all three curriculum bands achievable (${\geq}2$ unique models
each). $^\dagger$CelebA's 166 high-$J$ models are repeated evaluations
of a single architecture (\texttt{ga-352}). $^\ddagger$SVHN reaches
$J{=}0.82$ with a \texttt{unq}-family anchor but lacks high and medium
bands; only L3-equivalent generation is possible.}
\label{tab:viability}
\end{table}

CIFAR-10 is the only fully viable dataset: its 13,023 procedurally
generated \texttt{rl-bb-init} architectures span $J{=}0.30$--$0.98$,
enabling all three curriculum bands. These are properties of the
current database state, not of the method itself.

\subsubsection{Band Construction and Curriculum Schedule}
 
Similarity bands are defined as named ranges in \texttt{SIM\_BANDS} (\texttt{Query.py}) and enforced directly in the SQL WHERE clause via the \texttt{jaccard\_blobs} UDF. Five bands are defined; the intermediate low band ($J = 0.60$--$0.85$) is merged with the very-low-near band to form a unified Level~3 band spanning $[0.30, 0.85)$, providing a richer retrieval pool without requiring extreme structural contrasts:
 
\begin{itemize}
    \item \textbf{L1 — High similarity:} $J \in [0.95, 0.98)$, $k = 2$
    \item \textbf{L2 — Medium similarity:} $J \in [0.85, 0.95)$, $k = 2$ then $k = 3$
    \item \textbf{L3 — Merged low / very-low-near:} $J \in [0.30, 0.85)$, $k = 2$, $k = 3$, and $k = 4$
\end{itemize}

\subsubsection{Automatic Anchor Selection}
 
With 13{,}000+ models it is impractical to supply \texttt{anchor\_nn} manually. \texttt{\_resolve\_anchor()} (\texttt{Query.py}) selects the anchor automatically: it iterates candidate architectures in descending accuracy order and selects the first model that has at least $k$ neighbours in the requested similarity band, verified by running the \texttt{jaccard\_blobs} UDF via a COUNT subquery. The \texttt{anchor\_nn} parameter in \texttt{JoinConf} is therefore optional; when absent, the auto-selected anchor is recorded in \texttt{curriculum\_meta} alongside \texttt{auto\_anchor: true} for reproducibility. All curriculum experiments use anchor \texttt{rl-bb-init-9afa16ef}, selected automatically as the highest-accuracy CIFAR-10 architecture with sufficient neighbours across all three similarity bands.
 
\subsection{Prompt Design Across Curriculum Levels}
 
Generation uses a constrained XML-style interface with three mandatory output regions: neural architecture code, hyperparameter specification, and data transform definition. Prompts are constructed by \texttt{NNGenPromptCurriculum.py} in TALL selection mode: the prompt contains the anchor architecture as primary reference and $k$ nearest neighbours retrieved via the anchor-band UDF query. Constraints are tightened across levels in response to observed failure modes:
 
\begin{itemize}
    \item \textbf{Level 1} prompts emphasise stable module scaffolding and forward-pass structure, addressing the dominant L1 failure of incorrect constructor declarations.
    \item \textbf{Level 2} prompts add backbone usage restrictions and variable naming consistency.
    \item \textbf{Level 3} prompts add dimension-inference and feature-map routing constraints to address tensor shape mismatches that first appear at this diversity level.
\end{itemize}
 
Proven CIFAR-10 prompts are stored as JSON configuration files (\texttt{Curriculum\_L*.json}) and adapted for cross-dataset experiments by substituting only the dataset name, transform, input size, and architecture-specific constraints. The anchor selection and band-filtered reference retrieval operate identically across datasets.

\subsection{Sequential Adapter Merging}
\label{sec:merging_analysis}

For level $t$, we select the best epoch by composite score:
\[
S_t = \frac{|V_t|}{N} \times \max_{i \in V_t} a_i,
\]
where $V_t$ is the evaluator-valid generation set. We then merge the
corresponding adapter into the backbone before the next level.

\paragraph{Quantization and precision.}
OlympicCoder-7B is loaded in 4-bit NF4 during generation and
fine-tuning; adapters are maintained in bfloat16. At each merge step,
base weights are dequantized to float16, the adapter is applied as
$W_{\text{merged}} = W_{\text{base}} + \Delta W$, and the result saved
as float16. Subsequent levels re-quantize to NF4 on load. This
dequantize--merge--requantize cycle introduces ${\approx}0.1$--$0.3\%$
relative quantization error per step. The procedure is identical across
all experimental conditions.

Direct weight-space comparison between the 4-bit base and float16
merged model is uninformative due to the format difference; behavioural
evidence  evaluator acceptance rates across levels  serves as the
primary indicator of whether earlier coding conventions are retained.

\section{Experiments}

\subsection{Experimental Setup}

All experiments were conducted on a single NVIDIA GeForce RTX~4090
(24\,GB VRAM). Both LLM adaptation and CNN evaluation share the GPU,
motivating the sequential evaluation design.

\paragraph{Base language model.}
OlympicCoder-7B~\cite{OlympicCoder2025}: Qwen2-based, 7B
parameters, 32k context window. Loaded in 4-bit NF4; fine-tuning in
bfloat16.

\paragraph{LoRA configuration.}
Rank $r{=}32$, $\alpha{=}32$, dropout $0.05$, applied to all attention
and MLP projections across transformer layers 0--23.
\texttt{paged\_adamw\_8bit}, learning rate $10^{-6}$, batch size $1$,
gradient accumulation 4 steps, cosine scheduling.

\paragraph{Generation.}
$N{=}15$ candidates per epoch; temperature $1.0$, top-$k{=}50$,
top-$p{=}0.9$, maximum 16,384 decoding tokens. Candidates are
submitted directly to the LEMUR evaluator. Primary experiments use no
post-processing repair; the $2{\times}2$ ablation additionally tests a
two-step partial repair (injecting missing
\texttt{supported\_hyperparameters()} and \texttt{train\_setup()}
declarations when absent).

\paragraph{Evaluator-side training.}
CIFAR-10 classification under \texttt{norm\_256\_flip}, one training
epoch, SGD with LLM-proposed hyperparameters, batch size capped at 16,
8-minute execution limit per candidate. We retain the sequential
\texttt{NNEval.py} path over the multiprocessing worker-pool, which
introduced GPU memory conflicts on a single RTX~4090.

\subsection{Evaluation Criteria}

\paragraph{Composite score.}
\begin{equation}
S = \frac{|V|}{N} \times \max_{i \in V} a_i, \quad S{=}0
\text{ if } V{=}\emptyset.
\label{eq:score}
\end{equation}
This jointly captures search efficiency (via SR $= |V|/N$) and
architectural quality (via peak accuracy). We additionally report mean
accuracy of valid generations. With $N{=}15$, differences in composite
score below $0.05$ are treated as within-noise given the small sample
size. Wilson 95\% confidence intervals are reported for SR; sliding-window
$t$-distribution CIs for composite scores and accuracy trajectories.

\paragraph{Statistical scope.}
All curriculum results are from a single experimental seed (seed~2).
We do not report multi-seed means or significance tests, as the
computational budget permits only one trajectory per condition. Results
should be interpreted as single-run observations rather than
statistically validated findings.

\section{Results}
\label{sec:results}

\subsection{Curriculum Progression on CIFAR-10}

\begin{figure*}[!bt]
    \centering
    \includegraphics[width=0.49\linewidth]{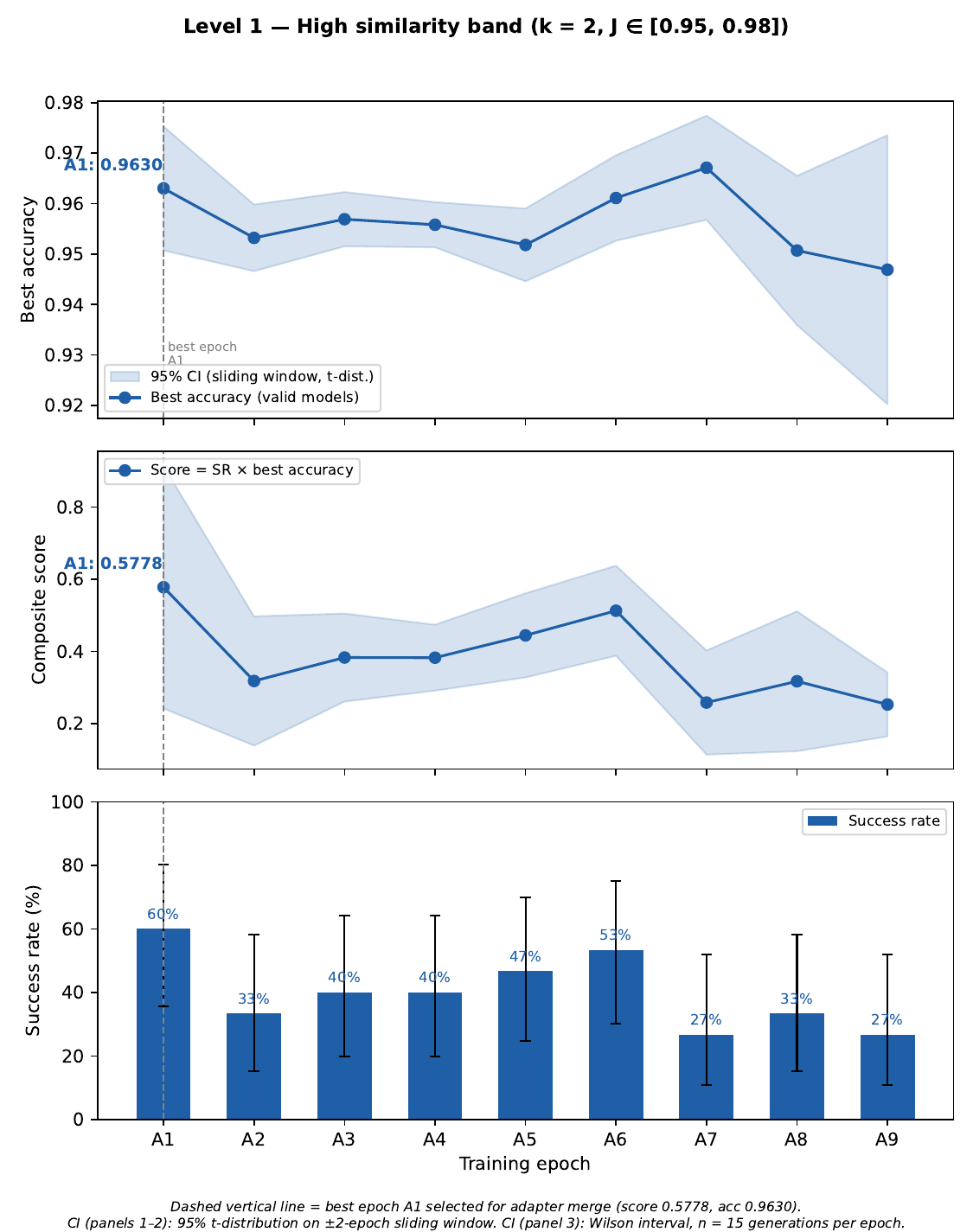}
    \includegraphics[width=0.49\linewidth]{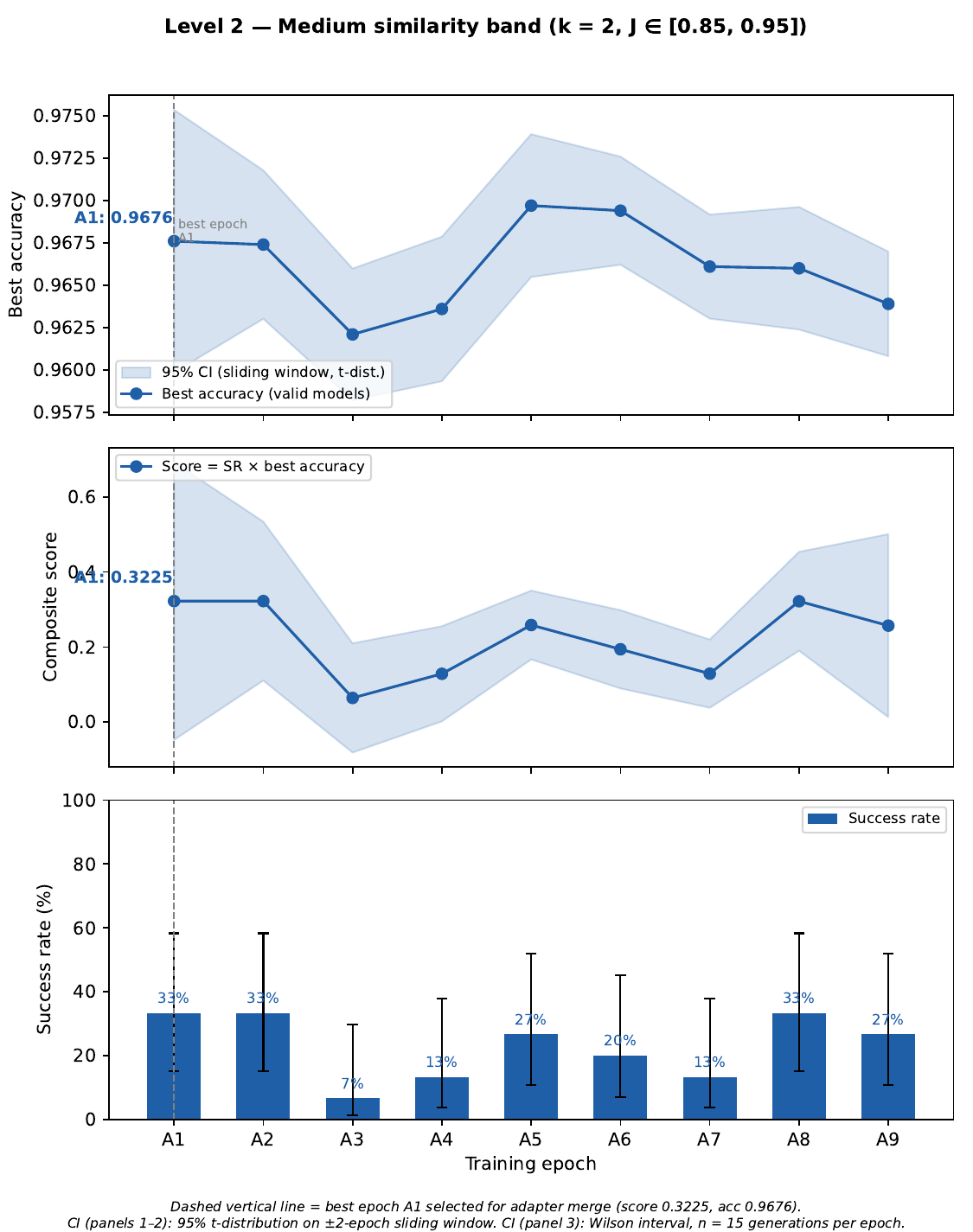}
    \caption{Level~1 $k{=}2$ (left) and Level~2 $k{=}2$ (right)
    per-epoch trajectories. Top: accuracy with 95\% CI. Middle: SR
    with Wilson 95\% CI. Bottom: composite score with sliding-window CI.
    Dashed vertical line = epoch selected for adapter merge.}
    \label{fig:collage1}
\end{figure*}

\begin{figure*}[!bt]
    \centering
    \includegraphics[width=0.49\linewidth]{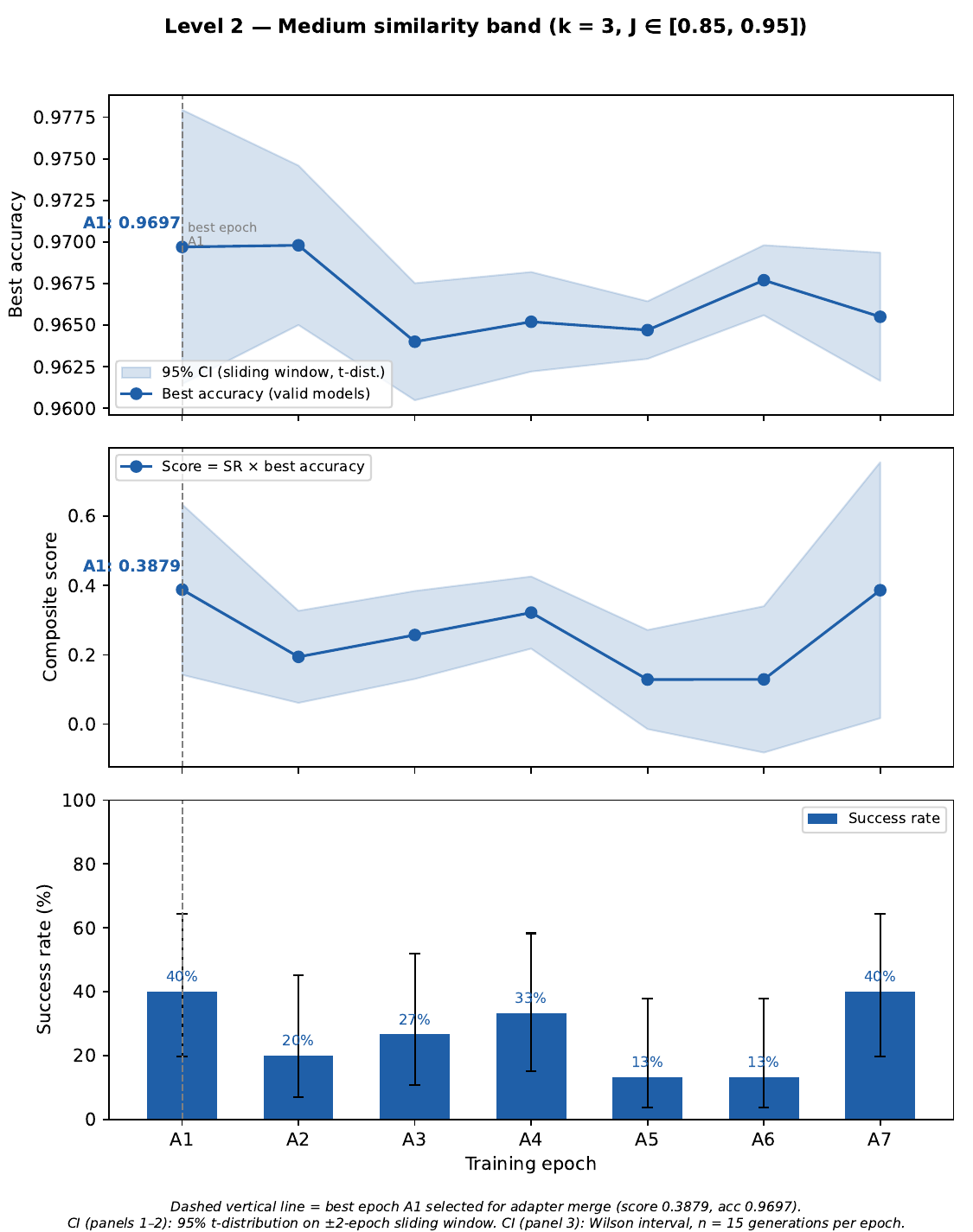}
    \includegraphics[width=0.49\linewidth]{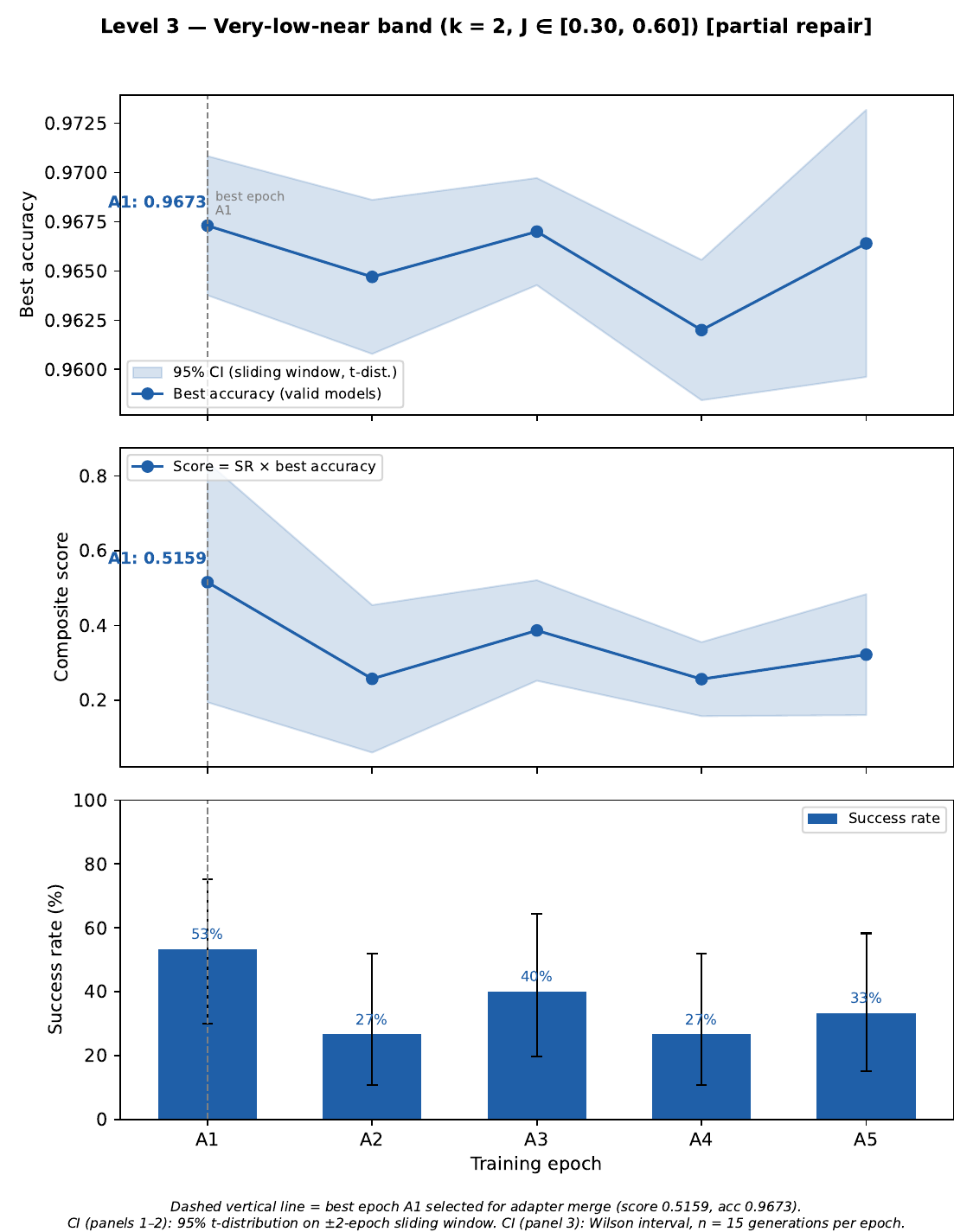}
    \caption{Level~2 $k{=}3$ (left) and Level~3 $k{=}2$ (right).
    L3~$k{=}2$ uses partial repair; dashed vertical line = merge epoch.}
    \label{fig:collage2}
\end{figure*}

\begin{figure*}[!bt]
    \centering
    \includegraphics[width=0.49\linewidth]{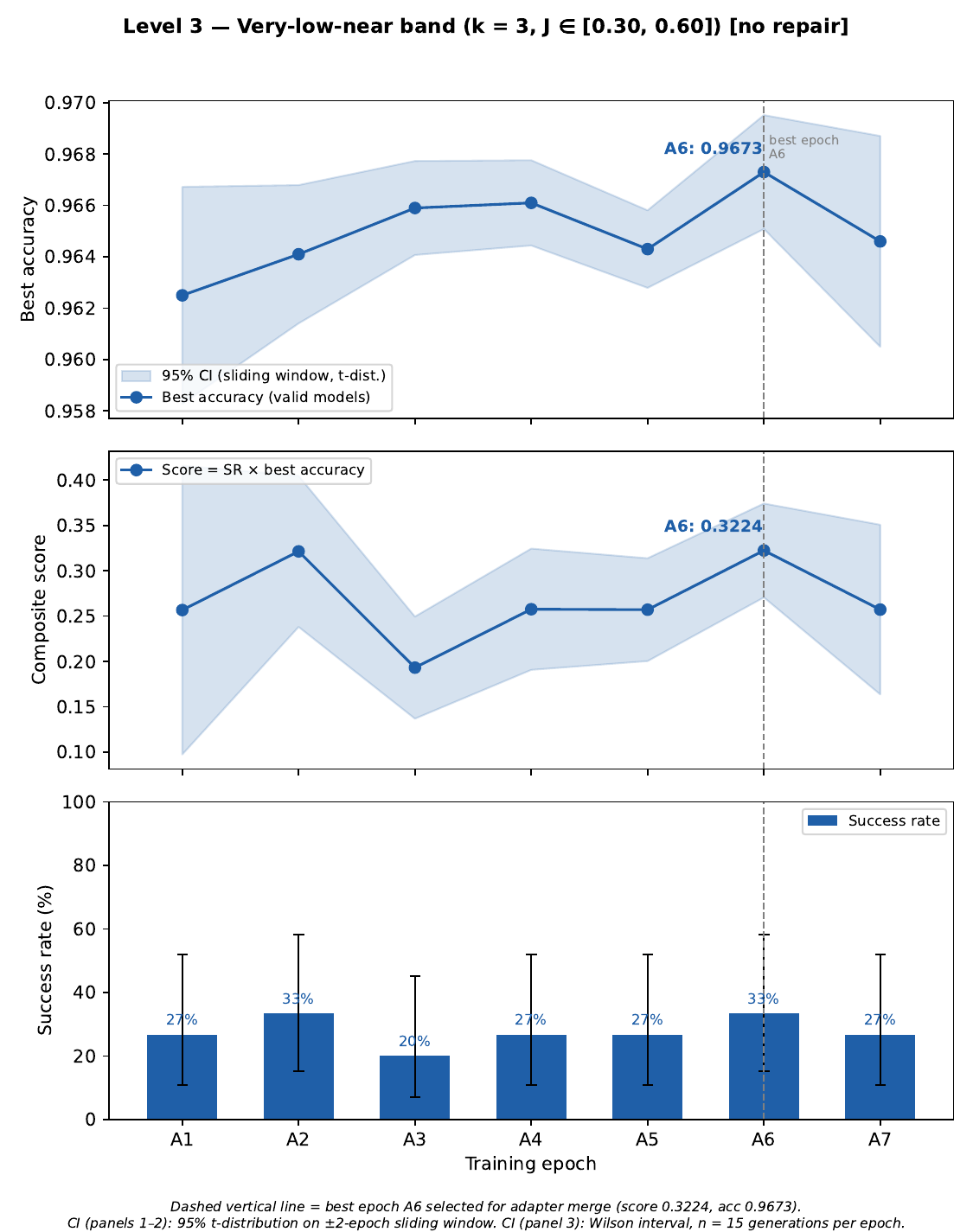}
    \includegraphics[width=0.49\linewidth]{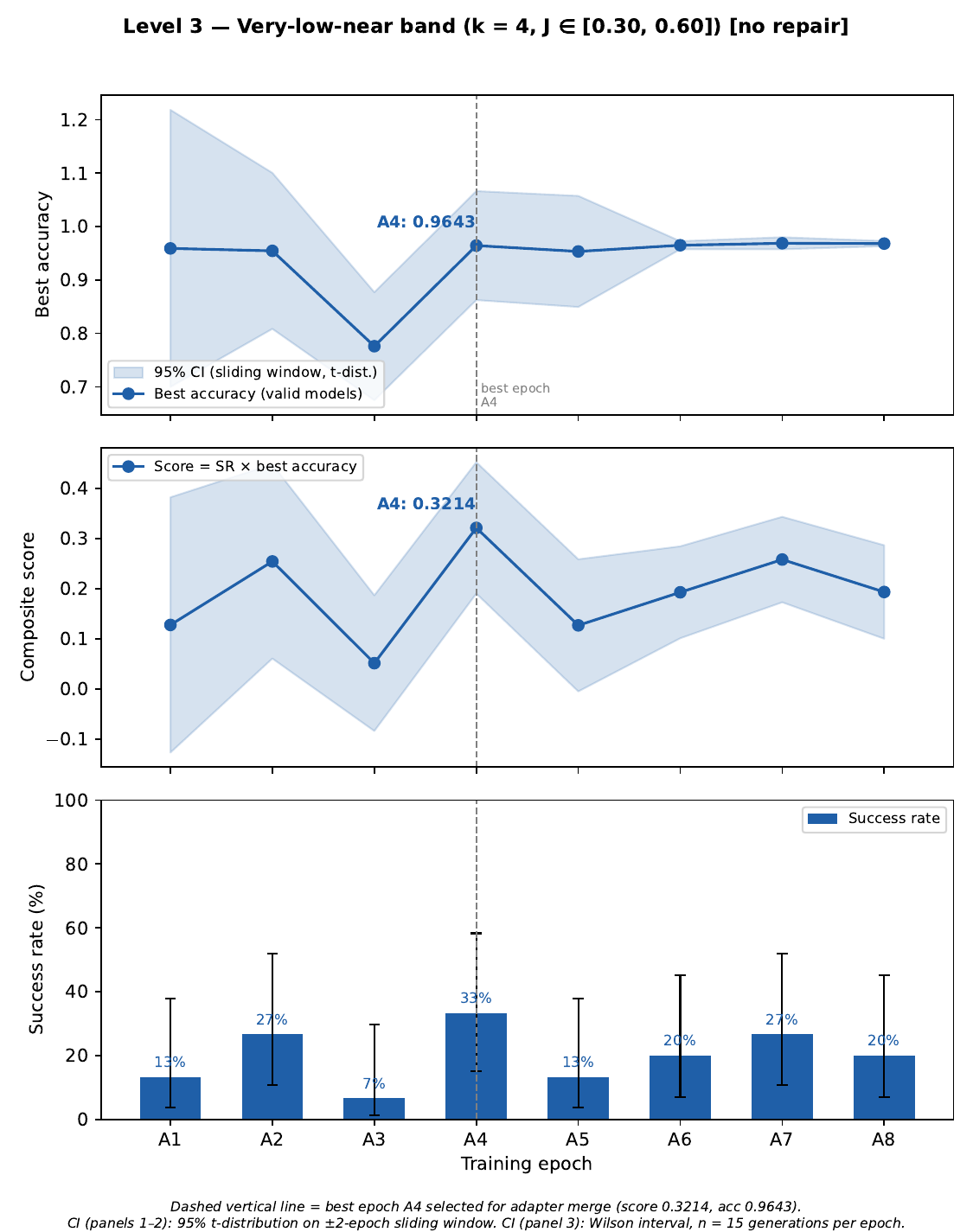}
    \caption{Level~3 $k{=}3$ (left) and Level~3 $k{=}4$ (right).
    Both without repair; dashed vertical line = merge epoch.}
    \label{fig:collage3}
\end{figure*}

Table~\ref{table:curriculum} reports best-epoch results per curriculum
level; per-epoch trajectories are shown in
Figures~\ref{fig:collage1}--\ref{fig:collage3}.

\begin{table*}[t]
    \centering
    \fontsize{7.5}{9}\selectfont
    \begin{tabular}{l c c c c c c c c}
        \toprule
        \textbf{Level} & \textbf{Band} & \textbf{$k$} &
        \textbf{Epochs} & \textbf{Best ep.} &
        \textbf{Best acc.} & \textbf{Best SR} &
        \textbf{Mean SR} & \textbf{Score} \\
        \midrule
        L1 & High ($0.95$--$0.98$)        & 2 & 9 &
            A1  & 0.9630 & 60\% & 40\% & 0.578 \\
        L2 & Medium ($0.85$--$0.95$)      & 2 & 9 &
            A1  & 0.9676 & 33\% & 23\% & 0.323 \\
        L2 & Medium ($0.85$--$0.95$)      & 3 & 7 &
            A1  & 0.9697 & 40\% & 27\% & 0.388 \\
        L3 & Low/VL-near ($0.30$--$0.85$) & 2 & 4 &
            A1$^*$ & 0.9673 & 53\% & 37\% & 0.516 \\
        L3 & Low/VL-near ($0.30$--$0.85$) & 3 & 8 &
            A2/A6 & 0.9673 & 33\% & 25\% & 0.322 \\
        L3 & Low/VL-near ($0.30$--$0.85$) & 4 & 8 &
            A4  & 0.9643 & 33\% & 20\% & 0.321 \\
        \bottomrule
    \end{tabular}
    \caption{Best-epoch results per curriculum level on CIFAR-10,
    $N{=}15$, single seed, no post-processing repair unless noted.
    $^*$L3~$k{=}2$ uses two-step partial repair (Section~\ref{sec:2x2}).
    Score $=$ SR $\times$ best accuracy (Eq.~\ref{eq:score}).
    Adapters merged cumulatively from OlympicCoder-7B.}
    \label{table:curriculum}
\end{table*}

\paragraph{Level~1.}
With references constrained to $J{\in}[0.95,0.98)$, both references
belong to the same \texttt{rl-bb-init} family. The L1 prompt
emphasises evaluator interface internalisation  correct
hyperparameter declaration and \texttt{learn()} structure  rather
than architectural diversity. Without any repair, we observe $60\%$ SR
at epoch~A1 (score $0.578$, accuracy $96.30\%$) and $40\%$ mean SR
across 9 epochs. This establishes that evaluator-compatible generation
from high-similarity references is feasible without post-processing
intervention.

\paragraph{Level~2.}
Moving to $J{\in}[0.85,0.95)$ after the L1 merge, the prompt adds
backbone usage restrictions. At $k{=}2$, SR is $33\%$ best and $23\%$
mean over 9 epochs. At $k{=}3$, we observe $40\%$ best SR at A1
(score $0.388$, accuracy $96.97\%$) with $27\%$ mean SR over 7 epochs.
Accuracy of valid generations at both $k$ values ($96.2$--$96.97\%$)
remains high relative to the reference accuracy ceiling ($97.17\%$),
suggesting L1 coding priors persist into the medium-diversity stage.

\paragraph{Level~3.}
At L3 the curriculum first encounters the \texttt{rl-back-init}
architecture family, which uses qualitatively different training loop
conventions. Without repair, the first three L3~$k{=}2$ epochs produce
$7\%$ SR ($1/15$ valid per epoch). Error analysis of epoch~A1 shows
that $40\%$ of the 14 failures per epoch are interface-level  missing
\texttt{train\_setup} or absent \texttt{supported\_hyperparameters} 
consistent with progressive forgetting of L1 boilerplate across merge
steps, though alternative explanations (prompt ambiguity, reference
family shift) cannot be excluded from behavioural evidence alone.

The $k{=}3$ and $k{=}4$ configurations, run without repair from the
merged model, stabilise at $27$--$33\%$ SR (best scores $0.322$ and
$0.321$). The $k{=}4$ run shows a collapse at epoch~A3 (accuracy
$77.6\%$, SR $7\%$) attributable to a pathologically misconfigured
hyperparameter generation, before recovering to $33\%$ SR at A4. The
scores at $k{=}3$ and $k{=}4$ are statistically indistinguishable
given $N{=}15$.

\subsection{Repair--Curriculum Interaction}
\label{sec:2x2}

Table~\ref{tab:2x2} presents the $2{\times}2$ ablation at Level~3
$k{=}2$. All conditions use the same reference pool, prompts, and
generation budget ($N{=}15$, single seed per condition).

\begin{table}[t]
    \centering
    \fontsize{7.5}{9}\selectfont
    \begin{tabular}{l c c c c}
        \toprule
        \textbf{Model} & \textbf{Repair} &
        \textbf{Best SR} & \textbf{Mean SR} & \textbf{Score} \\
        \midrule
        Base       & \xmark & 47\% & 26\% & 0.450 \\
        Curriculum & \xmark &  7\% &  7\% & 0.064 \\
        Base       & \cmark & 53\% & -- & 0.516 \\
        Curriculum & \cmark & 53\% & 37\% & 0.516 \\
        \bottomrule
    \end{tabular}
    \caption{$2{\times}2$ ablation at Level~3 $k{=}2$ on CIFAR-10.
    Repair = two-step partial \texttt{fix\_param\_usage}.
    All conditions: $N{=}15$, identical references, single seed.
    Mean SR for Base+repair is not reported (single epoch observed).}
    \label{tab:2x2}
\end{table}

Without repair, the base model ($47\%$ best SR, $26\%$ mean SR)
substantially outperforms the curriculum model ($7\%$ SR across all
epochs). This pattern is consistent with merge-level drift: the three
sequential merge steps preceding L3 may shift backbone weights away
from the evaluator interface patterns learned at L1. The base model,
whose weights have not undergone this drift, retains the original
interface knowledge. These observations are behavioural; the underlying
weight-space changes cannot be directly measured due to the 4-bit
NF4/float16 format difference between base and merged models.

With partial repair, both conditions reach $53\%$ best SR and score
$0.516$. Repair equalises the two models in this single-seed
comparison. The finding that curriculum and base model perform
equivalently under repair suggests that, within this configuration,
the accumulated L1/L2 architectural priors provide no measurable
advantage at Level~3 once the interface problem is addressed. Whether
this holds across seeds, different diversity levels, or different base
LLMs is an open question that multi-seed experiments would need to
address.

\subsection{Structural Similarity Between Generated and Reference Architectures}

To examine whether the curriculum produces genuine architectural synthesis
rather than direct reference reproduction, the structural similarity between
each generated model and the fixed anchor architecture is measured. The
anchor, \texttt{rl-bb-init-9afa16ef}, is the same highest-accuracy CIFAR-10
architecture used to define the similarity bands throughout curriculum
construction, providing a stable structural reference frame across all
levels.

Similarity is computed using the same MinHash representation employed during
curriculum construction: 128-permutation signatures over 7-gram shingles
extracted from tokenised source code. Using a consistent representation for
both curriculum definition and post-training analysis allows the generated
architectures to be evaluated within the same structural space used to
organise the curriculum. Table~\ref{tab:structural-similarity} summarises
the mean and maximum Jaccard similarity between generated models and the
anchor, together with the mean downstream accuracy at each curriculum level.

\paragraph{Generated architectures remain structurally distinct from the
anchor, and increasingly so as the curriculum progresses.} At Level~1, where
training references are drawn from the high-similarity band relative to the
anchor, generated architectures achieve a mean similarity of $0.513$ to the
anchor itself. This indicates that the model does not simply reproduce the
anchor architecture even when trained on its closest neighbours; instead,
generated architectures occupy a related but structurally distinct position.
A similar pattern holds at Level~2, where medium-band training yields
generations with mean anchor similarity of $0.461$ ($k{=}2$) and $0.450$
($k{=}3$).

\paragraph{Similarity to the anchor decreases across the curriculum.} Mean
similarity between generated architectures and the anchor decreases from
$0.513$ at Level~1 to between $0.236$ and $0.266$ at Level~3. The largest
reduction occurs during the transition from the medium-similarity stage to
the very-low-near stage ($0.450 \rightarrow 0.266$), corresponding to the
point at which the curriculum introduces structurally distant reference
families. Within Level~3 itself, similarity to the anchor is essentially
flat across $k{=}2$, $k{=}3$, and $k{=}4$ ($0.266$, $0.236$, $0.236$),
indicating that once the curriculum reaches the very-low-near band, further
increases in reference-set diversity ($k$) do not produce additional
anchor-relative divergence in the generations themselves. The overall
decline from Level~1 to Level~3 is nonetheless substantially steeper than
the corresponding gap between training references at each band, suggesting
that the curriculum does not merely cause generations to mirror the
diversity already present in the reference set, but actively shifts the
generator toward progressively less anchor-dependent outputs as training
references diversify.

\paragraph{Structural novelty does not reduce downstream performance.} The
lowest similarity values are observed at Level~3, where mean accuracy
remains high and stable: $94.97\%$ at $k{=}2$, $93.19\%$ at $k{=}3$, and
$94.18\%$ at $k{=}4$. Despite producing the most structurally distinct
generations relative to the anchor, with mean similarity less than half
that of Level~1, Level~3 models match or exceed the mean accuracy achieved
at Levels~1 and~2 ($93.40\%$ and $94.27$--$94.58\%$ respectively). This
indicates that the curriculum does not force a trade-off between
architectural novelty and evaluator performance: as the curriculum
progresses, the model generates architectures that diverge substantially
from the anchor while still preserving the structural properties necessary
for strong CIFAR-10 performance.

\paragraph{Consistent anchor selection across curriculum levels.} All
curriculum stages use the same anchor model, \texttt{rl-bb-init-9afa16ef},
selected automatically by the retrieval procedure as the highest-accuracy
CIFAR-10 architecture in the LEMUR database with sufficient neighbours
across all similarity bands. Using a fixed anchor across levels provides a
stable structural reference frame throughout the curriculum, so changes in
similarity statistics reflect differences in reference diversity and
generator behaviour rather than changes in the local neighbourhood of the
anchor itself.

\begin{table*}[h]
   \centering
   \begin{tabular}{llccccc}
   \toprule
Level & Band & $k$ & Mean $j$ & Max $j$ & Mean Acc. & $N$ \\
\midrule
L1 & High            & 2 & 0.513 & 0.648 & 93.40\% & 110 \\
L2 & Medium          & 2 & 0.461 & 0.484 & 94.27\% & 85  \\
L2 & Medium          & 3 & 0.450 & 0.484 & 94.58\% & 84  \\
L3 & Very-low-near   & 2 & 0.266 & 0.484 & 94.97\% & 225 \\
L3 & Very-low-near   & 3 & 0.236 & 0.258 & 93.19\% & 62  \\
L3 & Very-low-near   & 4 & 0.236 & 0.250 & 94.18\% & 67  \\
 \bottomrule
 \end{tabular}
 \caption{Mean and maximum Jaccard similarity between generated architectures
and the fixed anchor model, together with mean generation accuracy and the
number of evaluated generations ($N$) per level. Similarity is computed
using 128-permutation MinHash signatures over 7-gram code shingles. The
anchor model is \texttt{rl-bb-init-9afa16ef} across all curriculum levels.}
\label{tab:structural-similarity}
\end{table*}

\begin{figure}[h]
    \centering
    \includegraphics[width=\linewidth]{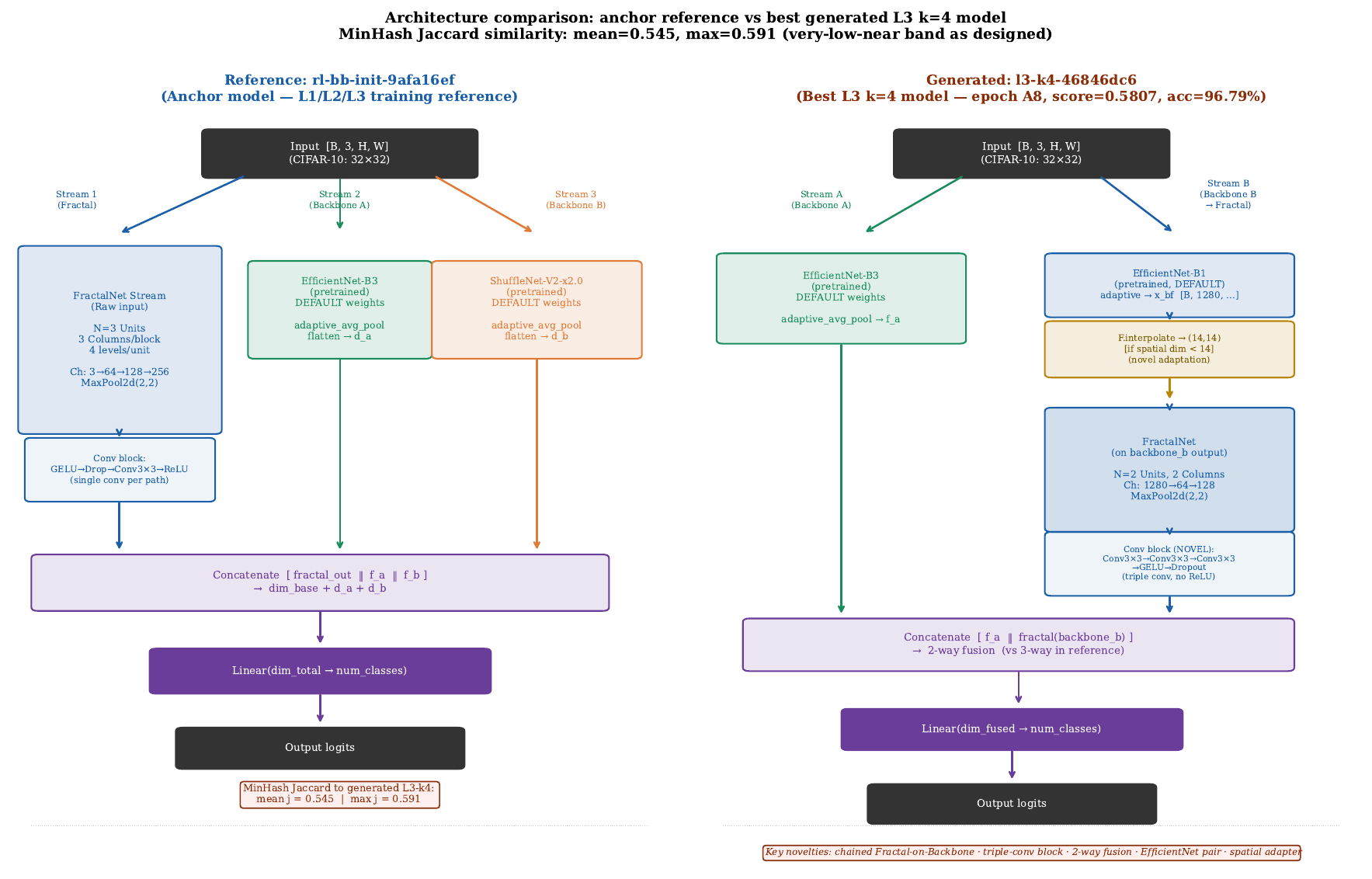}
    \caption{Architecture comparison between the anchor reference model
    (left) and the best generated model at Level~3 $k{=}4$ (right).
    Both share the FractalNet motif and pretrained backbone integration,
    but differ in fusion topology (3-way parallel vs.\ 2-way chained),
    conv block design (single vs.\ triple convolution), backbone pair
    selection, and spatial resolution adaptation.
    MinHash Jaccard: $\bar{j}=0.545$ (very-low-near band).}
    \label{fig:arch_comparison}
\end{figure}

\paragraph{Case study: anchor reference and generated architecture.}
Figure~\ref{fig:arch_comparison} compares the anchor reference model,
\texttt{rl-bb-init-9afa16ef}, with the highest-scoring architecture
generated at Level~3 with $k{=}4$, \texttt{l3-k4-46846dc6}
(epoch~A8, score $0.581$, accuracy $96.79\%$). The generated model
achieves a mean MinHash Jaccard similarity of $\bar{j}=0.545$ relative
to its reference set, placing it within the intended very-low-near
band.

The two architectures share several high-level design elements
associated with the \texttt{rl-bb-init} family: both combine a
FractalNet-style module with pretrained TorchVision backbones and merge
the resulting feature streams through concatenation before the final
classifier. They also retain common implementation patterns, including
the \texttt{FractalUnit} structure, \texttt{MaxPool2d(2,2)}
downsampling, adaptive pooling and flattening utilities, gradient
clipping at $3.0$, and the required LEMUR interface functions.

The generated model introduces four structural changes not present in
the training references. First, whereas the reference uses a
three-branch parallel fusion design, the generated architecture adopts
a partially sequential topology: features extracted by
\texttt{backbone\_b} (EfficientNet-B1) are passed into the FractalNet
module rather than processed independently. Second, the convolutional
blocks differ: the reference uses a single convolution layer with GELU,
dropout, and ReLU, while the generated model employs three consecutive
convolution layers followed by GELU and dropout without ReLU. Third,
\texttt{ShuffleNet-V2-x2.0} is replaced by a second EfficientNet-B1
branch, creating a homogeneous backbone pairing. Fourth, an explicit
spatial resizing step using \texttt{F.interpolate} to enforce a minimum
feature resolution of $14 \times 14$ is introduced --- a robustness
mechanism absent from all training references.

Overall, the generated architecture preserves the broader design
vocabulary of the reference family while introducing nontrivial
changes in topology, backbone selection, and feature routing. The model
recombines learned architectural components into new configurations
rather than reproducing the training references directly.

\subsection{Cross-Dataset Transfer: SVHN}

To probe whether base-model generation reliability transfers across
datasets, we run one single-level generation experiment on SVHN using
the \texttt{unq}-family anchor (\texttt{unq-af2a3d5f}), which achieves
$J{=}0.31$--$0.82$ similarity coverage within the SVHN database 
sufficient for L3-equivalent generation but not for a full curriculum
(Table~\ref{tab:viability}).

\begin{table}[h]
    \centering
    \fontsize{7.5}{9}\selectfont
    \begin{tabular}{l c c c c}
        \toprule
        \textbf{Dataset} & \textbf{Model} & \textbf{Best SR} &
        \textbf{Mean SR} & \textbf{Best acc.} \\
        \midrule
        CIFAR-10 & Base, no repair & 47\% & 26\% & 0.9651 \\
        SVHN     & Base, no repair & 27\% & 10\% & 0.6045 \\
        \bottomrule
    \end{tabular}
    \caption{Cross-dataset comparison at L3-equivalent generation.
    CIFAR-10: \texttt{rl-bb-init} anchor, 7 epochs.
    SVHN: \texttt{unq}-family anchor, 9 epochs.
    Datasets, anchor families, and reference distributions differ;
    this comparison is observational, not controlled.}
    \label{tab:svhn}
\end{table}

SVHN generation yields substantially lower SR ($27\%$ best, $10\%$
mean) and best accuracy ($60.5\%$) compared to the CIFAR-10 base-model
condition ($47\%$ best SR, $96.5\%$ accuracy). Three collapse epochs
(SR $= 0\%$) suggest the \texttt{unq} architecture family presents a
harder synthesis target than \texttt{rl-bb-init}. Since the two
conditions differ in dataset, anchor family, and reference distribution,
we cannot attribute the difference to any single factor; we report this
as a preliminary cross-dataset transfer observation rather than a
controlled finding.

\subsection{Ordering vs.\ Random Baseline}

To directly examine whether ordering is the active variable, Table~\ref{tab:ordering-vs-random} compares the curriculum against a randomised ordering of the identical reference pool, the same base model, the same prompts, and the same sequential merging strategy. For seed 42, the curriculum steps were executed in the randomised order L3~$k{=}4$, L2~$k{=}2$, L3~$k{=}2$, L2~$k{=}3$, L3~$k{=}3$, with each step fine-tuning from the model produced by the previous step exactly as in the curriculum runs.

\paragraph{Random ordering matches curriculum peak performance at most steps.} At three of five steps, the best-epoch score under random ordering is statistically indistinguishable from the corresponding curriculum result: L3~$k{=}2$ reaches $0.516$ under random ordering versus $0.516$ under the curriculum, L2~$k{=}3$ reaches $0.387$ versus $0.388$, and L3~$k{=}3$ reaches $0.387$ versus $0.322$, with random ordering exceeding the curriculum result at this step. Only L2~$k{=}2$ shows a clear curriculum advantage at peak performance, with the curriculum reaching a best score of $0.322$ against $0.258$ for random ordering, a relative improvement of approximately $25\%$.

\paragraph{Random ordering is markedly less stable across epochs.} While peak performance is comparable, the per-epoch trajectories under random ordering show substantially higher variance and, in one case, complete collapse. At L2~$k{=}2$ under random ordering, epoch~7 produced zero successful models out of fifteen generations, a total failure not observed at any epoch in any curriculum run. Mean scores across all epochs of each step reflect this instability: random ordering's mean score at L2~$k{=}2$ is $0.142$, compared to a curriculum mean of $0.215$ at the same step, even though the gap in best-epoch score alone ($0.258$ vs.\ $0.322$) understates the difference once full-epoch reliability is taken into account.

\paragraph{Ordering affects reliability more than peak capability.} Taken together, these results suggest that the active benefit of progressive similarity-based ordering is not primarily an increase in the best architecture the model can reach within a step, since random ordering reaches comparable peaks at most steps tested. Instead, the curriculum's main advantage appears to be a reduction in epoch-to-epoch volatility and a lower risk of catastrophic single-epoch failure, particularly at the medium-similarity band. This reframes the curriculum's contribution from ``enables better generations'' to ``produces a more reliable fine-tuning trajectory,'' a distinction that matters for any practical deployment where a single bad epoch cannot simply be discarded after the fact.

\paragraph{Limitations of this comparison.} This comparison covers a single randomised ordering (seed 42) of the same five steps used in the curriculum; seeds 7 and 99 were not run due to time constraints, so the variance across different random orderings remains unquantified. A single seed cannot rule out the possibility that this particular random sequence was unusually favourable or unfavourable at any given step. The comparison also does not test alternative progressive orderings beyond the one used in the curriculum, such as interleaving $k$ values within each similarity level rather than completing all $k$ values at one level before advancing; whether such an alternative ordering would outperform either the curriculum or this random baseline remains an open question for future work.

\begin{table*}[t]
\centering

\begin{tabular}{lccccc}
\toprule
Step & \multicolumn{2}{c}{Best score} & \multicolumn{2}{c}{Mean score} & Best Acc. (rand.) \\
     & Curriculum & Random & Curriculum & Random & \\
\midrule
L2 $k{=}2$ & 0.322 & 0.258 & 0.215 & 0.142 & 0.9675 \\
L2 $k{=}3$ & 0.388 & 0.387 & 0.265 & 0.282 & 0.9676 \\
L3 $k{=}2$ & 0.516 & 0.516 & 0.368 & 0.322 & 0.9683 \\
L3 $k{=}3$ & 0.322 & 0.387 & 0.254 & 0.287 & 0.9680 \\
L3 $k{=}4$ & 0.33     & 0.387 & 0.20    & 0.161 & 0.9678 \\

\bottomrule

\end{tabular}
\caption{Best-epoch and mean-epoch composite score under curriculum ordering versus a single randomised ordering (seed 42) of the same five steps. Mean accuracy at the best epoch is also reported. Random ordering at L2~$k{=}2$ includes one epoch with zero successful generations (accuracy undefined), which is reflected in its lower mean score.}
\label{tab:ordering-vs-random}
\end{table*}

\subsection{Qualitative Comparison with Related Systems}

\begin{table*}[h]

    \centering
    \fontsize{7.5}{9}\selectfont
    \begin{tabular}{l c c c c c}
        \toprule
        \textbf{Method} & \textbf{FT} &
        \textbf{Curr.} & \textbf{Repair} &
        \textbf{1-GPU} & \textbf{NAS} \\
        \midrule
        OPRO~\cite{Yang2023OPRO}     & \xmark & \xmark & \xmark & \cmark & \xmark \\
        FunSearch~\cite{Romera2023FunSearch} & \xmark & \xmark & \xmark & \xmark & \xmark \\
        EvoPrompting~\cite{Chen2023EvoPrompting} & \xmark & \xmark & \xmark & \cmark & \cmark \\
        ReEvo~\cite{Ye2024ReEvo}     & \xmark & \xmark & \xmark & \xmark & \xmark \\
        NNGPT / LEMUR                & \cmark & \xmark & \xmark & \cmark & \cmark \\
        \textbf{This work}           & \cmark & \cmark & \cmark & \cmark & \cmark \\
        \bottomrule
    \end{tabular}
    \caption{Qualitative comparison. FT = LLM fine-tuning;
    Curr.\ = similarity-scheduled curriculum; 1-GPU = feasible on
    one consumer GPU; NAS = targets architecture search.}
    \label{tab:comparison}
\end{table*}

We are the only system in this family combining curriculum LLM
fine-tuning, single-GPU feasibility, and NAS targeting. A quantitative
comparison against a no-curriculum single-stage baseline and the
LEMUR-database best architecture is an important missing piece; this is
deferred to future work alongside multi-seed runs.

\subsection{Limitations}
\label{sec:limitations}

\paragraph{Statistical scope.}
All curriculum results are single-seed, single-model, single-anchor
observations with $N{=}15$ generations per epoch. No significance
tests are reported because the sample sizes and number of trajectories
do not support them. Differences in composite score $<0.05$ should not
be interpreted as meaningful. Multi-seed experiments with $N{\geq}30$
and paired significance testing (e.g., Wilcoxon signed-rank) are
required to validate the observed trends.

\paragraph{Single base model and anchor.}
All results use OlympicCoder-7B and anchor \texttt{rl-bb-init-9afa16ef}.
Generalisation to other base LLMs (e.g., Qwen2.5-Coder, DeepSeek-Coder)
and other anchors is entirely open. The curriculum effect observed here
may be specific to this model--anchor combination.

\paragraph{Repair confound at Level~3.}
The \texttt{fix\_param\_usage} repair module has a larger effect on
generation reliability than any curriculum manipulation at Level~3
(lifting SR from $7\%$ to $53\%$ for the curriculum model, and from
$47\%$ to $53\%$ for the base model). A complete characterisation
requires reporting the fraction of accepted generations that required
non-trivial repair, and running a repair-only baseline on a random
generator.

\paragraph{Random baseline incomplete.}
The central claim that ordering is the active ingredient rests on
Table~\ref{tab:ordering-vs-random}, which compares best-epoch and
mean-epoch composite scores under curriculum ordering against a single
randomised ordering (seed~42) of the same five steps. Because random
ordering matches curriculum peak performance at most steps tested,
running random ordering under two to three additional seeds is
necessary before the variance claim can be considered statistically
supported.

\paragraph{Evaluation budget.}
The 8-minute, single-epoch evaluation limit favours architectures
converging rapidly and limits conclusions about absolute accuracy.
Best-epoch model selection on the evaluation set inflates reported
scores; a held-out test set for final reporting would be preferable.

\paragraph{Dataset scope.}
CIFAR-10 is the only viable LEMUR dataset for full curriculum
construction (Table~\ref{tab:viability}). The SVHN cross-dataset
experiment is observational and not directly comparable to the CIFAR-10
conditions. The framework is dataset-agnostic and will extend naturally
as LEMUR database coverage grows.

\paragraph{Missing quantitative baseline.}
We do not report the accuracy of the best architecture in the LEMUR
database, a random-search baseline over the database, or the
unmodified OlympicCoder-7B without any fine-tuning, against which our
results should ultimately be positioned.

\section{Conclusion}
\label{sec:conclusion}

We have introduced a MinHash-based curriculum scheduling framework for
LLM-based neural architecture synthesis and reported four empirical
observations on CIFAR-10 with OlympicCoder-7B. First, progressive
similarity scheduling achieves $60\%$ peak SR at the high-similarity
level and $33$--$53\%$ at the most diverse level in single-seed runs
with $N{=}15$. Second, a $2{\times}2$ ablation at Level~3 shows that
without repair the base model ($47\%$ SR) substantially outperforms the
curriculum model ($7\%$ SR), while with partial repair both reach
$53\%$ SR  a pattern consistent with merge-level drift progressively
erasing interface priors. Third, structural similarity analysis
relative to the fixed anchor architecture shows that generated models
grow progressively less similar to the anchor as the curriculum
advances (mean Jaccard $j{=}0.513$ at Level~1 to $j{=}0.236$--$0.266$
at Level~3), while downstream accuracy remains stable or improves,
indicating that the curriculum drives genuine architectural divergence
from the anchor without sacrificing evaluator performance. Fourth,
SVHN base-model generation yields substantially lower SR and accuracy
than the CIFAR-10 equivalent, suggesting that cross-dataset transfer
of interface knowledge is imperfect and that the \texttt{unq}
architecture family presents a harder synthesis target.

These findings are preliminary: they rest on single-seed trajectories
with small generation batches, and the random-ordering comparison that
would directly validate the curriculum hypothesis is not yet complete.
Future work should address multi-seed statistical validation, extension
to additional base LLMs and anchors, completion of the random-ordering
baseline, and replacement of rule-based interface repair with a learned
correction mechanism.

{
  \small
  \bibliographystyle{IEEEtran}
  \bibliography{bibmain}
}

\end{document}